# Conceptual Text Summarizer:
# A New Model In Continuous Vector Space

**Mohammad Ebrahim Khademi*, Mohammad Fakhredanesh*, Seyed Mojtaba Hoseini***

** Faculty of Electrical and Computer Engineering, Malek Ashtar University of Technology, Iran*

## Abstract

Traditional methods of summarization are not cost-effective and possible today. Extractive summarization is a process that helps to extract the most important sentences from a text automatically and generates a short informative summary. In this work, we propose an unsupervised method to summarize Persian texts. This method is a novel hybrid approach that clusters the concepts of the text using deep learning and traditional statistical methods. First we produce a word embedding based on Hamshahri2 corpus and a dictionary of word frequencies. Then the proposed algorithm extracts the keywords of the document, clusters its concepts, and finally ranks the sentences to produce the summary. We evaluated the proposed method on Pasokh single-document corpus using the ROUGE evaluation measure. Without using any hand-crafted features, our proposed method achieves state-of-the-art results. We compared our unsupervised method with the best supervised Persian methods and we achieved an overall improvement of ROUGE-2 recall score of 7.5%.

# 1 Introduction

Automatic text summarization of a large corpus has been a source of concern over the years, from two areas of information retrieval and natural language processing. The primary studies in this field began in 1950s. Baxendale, Edmundson and Luhn have done research in those years [1]–[3]. Automatic generation of summaries provides a short version of documents to help users in capturing the important contents of the original documents in a tolerable time [4]. Now humans produce summaries of documents in the best way. Today, with the growth of data, especially in the big data domain, it is not possible to generate all of these summaries manually, because it's neither economical nor feasible.

There are two types of text summarization based on considering all or a specific part of a document:

- Generic summarization provides an overall summary of all information contained in a document. It answers the question "what is this document about?" A document is expected to contain several topics. Main topics are discussed extensively by many sentences. Minor topics have less sentence support and exist to support the main topics [5]. The specific goals of generic summarization are:

- To choose k number of sentences (as specified by the user) from a given document that best describe the main topics of the document.
  - To minimize redundancy of the chosen sentences.

- Query-relevant summarization is specific to information retrieval applications. It attempts to summarize the information a document contains pertaining to a specific search term [6]. The summary indicates “what this document says about the <query>”. The snippets bellow each result returned by a search engine is a common example for this type.

In addition there are two approaches to text summarization based on the chosen process of generating the summary [7]:

- Extractive summarization: This approach of summarization selects a subset of existing words, phrases, or sentences in the original text to form the summary. There are, of course, limitations on choosing these pieces. One of these limitations, which is common in summarization, is output summary length.
- Abstractive summarization: This approach builds an internal semantic representation and then uses natural language generation techniques to create a summary that is expected to be closer to what the text want to express.

Based on the current limitations of natural language processing methods, extractive approach is the dominant approach in this field. Almost all extractive summarization methods encounter two key problems in [8]:

- assigning scores to text pieces
- choosing a subset of the scored pieces

Hitherto text summarization has traveled a very unpaved path. In the beginning, frequency based approaches were utilized for text summarization. Then, lexical chain based approaches came to succeed with the blessing of using large lexical databases such as WordNet [9] and FarsNet [10], [11].

Hence, valid methods such as Latent Semantic Analysis (LSA) based approaches that do not use dedicated static sources -which requires trained human forces for producing them- became more prominent.

Word embedding models learn the continuous representation of words in a low-dimensional space [14], [17], [23]–[25]. In lexical semantics, Linear Dimension Reduction methods such as Latent Semantic Analysis have been widely used [14]. Non-linear models can be used to train word embedding models [26], [27]. Word embedding models not only have a better performance, but also lacks many problems of Linear Dimension Reduction methods such as Latent Semantic Analysis.

In this paper, a novel method of extractive generic document summarization based on perceiving the concepts present in sentences is proposed. Therefore after unsupervised learning of the target language word embedding, input document concepts are clustered based on the learned word feature vectors (hence the proposed method is language independent). After allocating scores to each conceptual cluster, sentences are ranked and selected based on the significance of the concepts present in each sentence. Ultimately we achieved promising results on Pasokh benchmark corpus.

The structure of the paper is as follows. Section two describes some related works. Section three presents the summary generation process. Section four outlines evaluation measures and experimental results. Section five concludes the paper and discusses the avenues for future research.

# 2 Related Works

Hitherto text summarization has traveled a very unpaved path. In the beginning, frequency based approaches were utilized for text summarization. Then, lexical chain based approaches came to succeed with the blessing of using large lexical databases such as WordNet [9] and FarsNet [10], [11]. Since the most common subject in the text has an important role in summarization, and lexical chain is a better criterion than word frequency for identifying the subject of text; as a result, a more discriminating diagnosis of the subject of text was made possible which was a further improvement in summarization. However the great reliance of these methods on lexical databases such as WordNet or FarsNet is the main weakness of these methods. For the success of these methods depends on enriching and keeping up to date the vocabulary of these databases that is very costly and time consuming, removing this weakness is not feasible.

Hence, valid methods such as Latent Semantic Analysis (LSA) based approaches that do not use dedicated static sources -which requires trained human forces for producing them- became more prominent. Latent Semantic Analysis is a valid unsupervised method for an implicit representation of the meaning of the text based on the co-occurence of words in the input document. This method is unsupervised and it is considered an advantage. But this method has many other problems:

- The dimensions of the matrix changes very often (new words are added very frequently and corpus changes in size).
- The matrix is extremely sparse since most words do not co-occur.
- The matrix is very high dimensional in general ( $\approx 10^6 \times 10^6$ )
- Quadratic cost to train (i.e. to perform SVD)

Many of Natural Language Processing (NLP) systems and methods consider words as separate units. In such systems, the similarity between words is not defined and words are considered as indexes of a dictionary. This approach is generally adopted for the following reasons:

- Simplicity
- Reliability
- Advantage of training large data volume over using complex models

As for the third reason, according to past observations and experiences, in general, simple models that are trained on a vast amount of data are more effective than complex models that are trained on quantitative data. Today we can assume that N-gram models can be trained on all existing data (billions of words [12]).

With the advent of machine learning methods in recent years, training more complex models on much larger datasets has become possible. lately, the advancement in computing power of GPUs and new processors have made it possible for hardwares to implement these more advanced models. One of the most successful of these cases in recent years is the use of the distributed representation of vocabularies [13].

Word Embedding model was developed by Bengio et al. more than a decade ago [14]. The word embedding model W, is a function that maps the words of a language into vectors with about 200 to 500 dimensions. To initialize W, random vectors are assigned to words. This model learns meaningful vectors for doing some tasks.

In lexical semantics, Linear Dimension Reduction methods such as Latent Semantic Analysis have been widely used [15]. Non-linear models can be used to train word embedding models [16], [17]. Word embedding models not only have a better performance, but also lacks many problems of Linear Dimension Reduction methods such as Latent Semantic Analysis.

Distributed representation of vocabularies (Word Embedding) is one of the important research topics in the field of natural language processing [13], [15]. This method, which in fact is one of the deep learning branches, has been widely used in various fields of natural language processing in recent years. Among these, we can mention the following:

- Neural language model [14], [18]
- Sequence tagging [19], [20]
- Machine translation [21], [22]
- Contrasting meaning [23]

Bengio et al. [14], Mikolov et al. [24], and Schwenk [18] have shown that Neural network based language models have produced much better results than N-gram models.

Although many text summarization methods are available for languages such as English, little work is done in devising methods of summarizing Persian texts.

In general these methods can be categorized as supervised and unsupervised, while most of the proposed methods so far have been of the former type. Supervised summarization methods presented for Persian documents are divided into four categories of heuristic, lexical chain based, graph based, and machine learning or mathematical based methods:

- Heuristic method:
    - Hassel and Mazdak proposed FarsiSum as a heuristic method [25]. It is one of the first attempts to create an automatic text summarization system for Persian. The system is implemented as a HTTP client/server application written in Perl. It has used modules

implemented in SweSum (Dalianis 2000), a Persian stop-list in Unicode format and a small set of heuristic rules.

- Lexical chain based methods:
  - Zamanifar et al. [26] proposed a new hybrid summarization technique that combined “term co-occurrence property” and “conceptually related feature” of Farsi language. They consider the relationship between words and use a synonym dataset to eliminate similar sentences. Their results show better performance in comparison with FarsiSum.
  - Shamsfard et al. [27] proposed Parsumist. They presented single-document and multi-document summarization methods using lexical chains and graphs. To rank and determine the most important sentence, they consider the highest similarity with other sentences, the title and keywords. They achieved better performance than FarsiSum.
  - Zamanifar and Kashefi [28] proposed AZOM, a summarization approach that combines statistical and conceptual text properties and in regards of document structure, extracts the summary of text. AZOM performes better than three common structured text summarizers (Fractal Yang, Flat Summary and Co-occurrence).
  - Shafiee and Shamsfard [29] proposed a single/multi-document summarizer using a novel clustering method to generate text summaries. It consists of three phases: First, a feature selection phase is employed. Then, FarsNet, a Persian WordNet, is utilized to extract the semantic information of words. Finally, the input sentences are clustered. Their proposed method is compared with three known available text summarization systems and techniques for Persian language. Their method obtains better results than FarsiSum, Parsumist and Ijaz.
- Graph based method:
  - Shakeri et al. [30] proposed an algorithm based on the graph theory to select the most important sentences of the document. They explain their objective as “The aim of this method is to consider the importance of sentences independently and at the same time the importance of the relationship between them. Thus, the sentences are selected to attend in the final summary contains more important subjects, and also have more contact with other sentences.” [30] Evaluation results indicate that the output of proposed method improves precision, recall and ROUGE-1 metrics in comparison with FarsiSum.
- Machine learning and mathematical based methods:
  - Kiyomarsi and Rahimi [31] proposed a new method for summarizing Persian texts based on features available in Persian language and the use of fuzzy logic. Their method obtains better results as compared with four previous methods.
  - Tofighy et al. [32] proposed a new method for Persian text summarization based on fractal theory whose main goal is using hierarchical structure of document to improve the summarization quality of Persian texts. Their method achieved a better performance than FarsiSum, but weaker than AZOM.

- Bazghandi et al. [33] proposed a textual summarization system based on sentence clustering. Collective intelligence algorithms are used for optimizing the methods. These methods rely on semantic aspect of words based on their relations in the text. Their results is comparable to traditional clustering approaches.
- Tofighi et al. [34] proposed an Analytical Hierarchy Process (AHP) technique for Persian text summarization. The proposed model uses the analytical hierarchy as a base factor for an evaluation algorithm. Their results show better performance in comparison with FarsiSum.
- Pourmasoumi et al. [35] proposed a Persian single-document summarization system called Ijaz. It is based on weighted least squares method [36]. Their results proved a better performance as compared with FarsiSum. They also proposed Pasokh [37], a popular corpus for evaluation of Persian text summarizers.

As an unsupervised method, Honarpisheh et al. [38] proposed a new multi-document multi-lingual text summarization method, based on singular value decomposition (SVD) and hierarchical clustering.

Success of Lexical chain based methods and supervised machine learning methods depends on enriching and keeping up to date lexical databases and training labeled datasets respectively, that is very costly and time consuming. These methods often use language-dependent features and can not be generalized to other languages. On the other hand unsupervised methods such as SVD based methods have many problems that are mentioned in the beginning of this section.

The proposed generic extractive method is a novel method that not only is unsupervised, but also does not have many problems of SVD-based methods and without using any hand-crafted features, achieves much better performance compared to supervised methods.

# 3 Proposed Algorithm

In this section we propose a novel method of extractive generic document summarization based on perceiving the concepts present in sentences. It is an unsupervised and language independent method that does not have many problems of SVD-based methods. For this purpose, firstly, the necessary preprocesses are performed on the Hamshahri2 corpus texts. Subsequently, the Persian word embedding is created by unsupervised learning of Hamshahri2 corpus. Then the input document keywords are extracted. Afterward the input document concepts are clustered based on the learned word feature vectors (hence the proposed method can be generalized to other languages), and the score of each of these conceptual clusters are calculated. Finally, the sentences are ranked and selected based on the significance of the concepts present in each sentence. The chart of this method is presented in Figure 1: Conceptual text summarizer. The following sections will be described based on this chart.

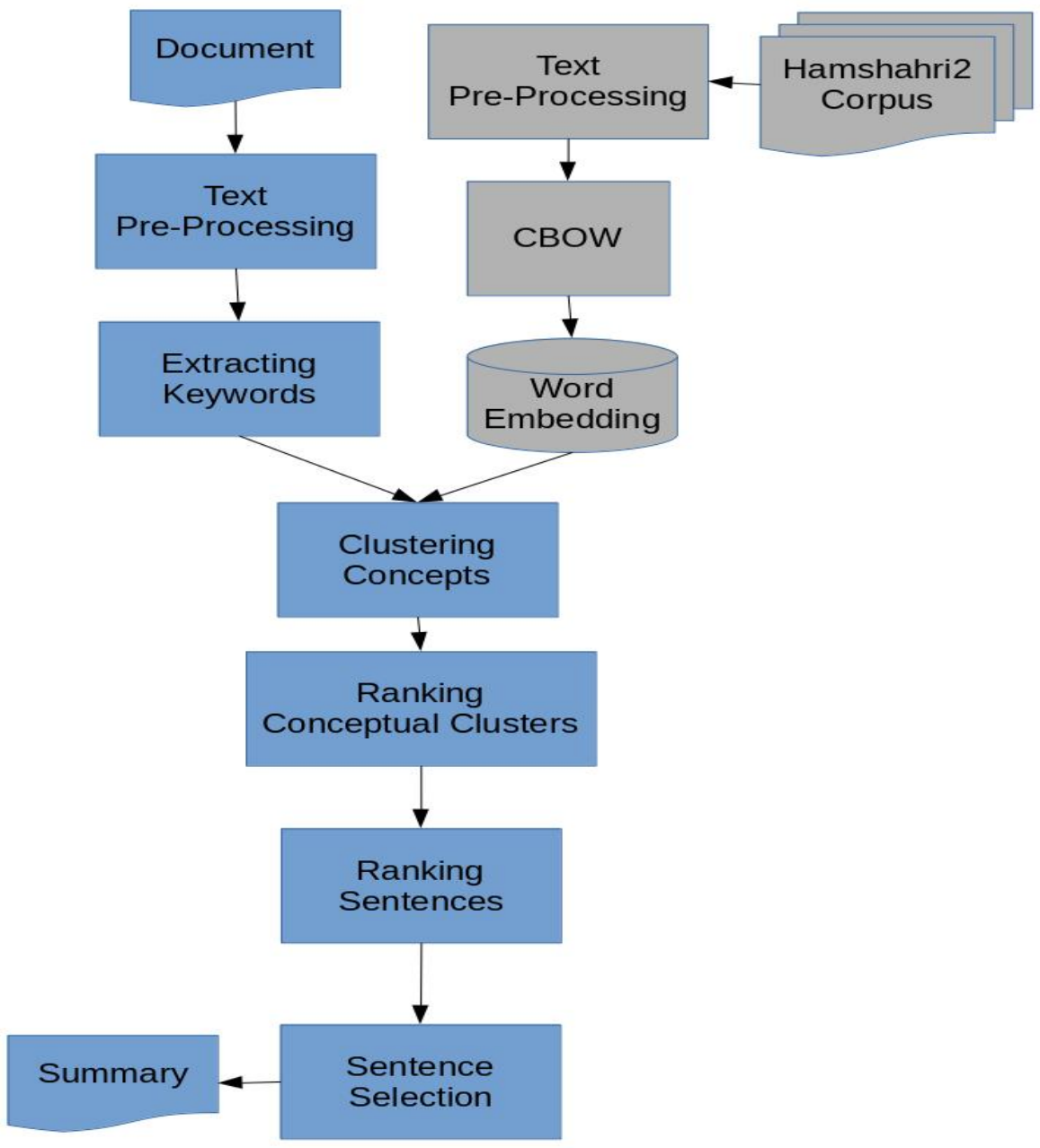

*Figure 1: Conceptual text summarizer*

## 3.1 Text Pre-Processing

To learn a Persian language model, we use Hamshahri2 [39] corpus. We need to produce a dictionary of vocabularies of Hamshahri2 corpus. To do this, we tokenize the words of each text file of the corpus using Hazm [40] library. Hazm is an applicable open source natural language processing library in Persian. Then we compose a dictionary out of these words by counting the frequency of each word throughout the corpus. This dictionary will be used in succeeding steps.

We constitute a complete list of Persian stopwords out of frequent words in the prepared dictionary along with stopword lists in other open source projects.

## 3.2 Unsupervised Learning of Persian Word Embedding

The Hamshahri2 corpus [39] has 3206 text files in unlabeled text sections. Each of these files is a concatenation of hundreds of news and articles. These news and articles are from different fields of cultural, political, social, etc.

To construct a suitable Persian word embedding set, we use CBOW model [41]. This model is a neural network with one hidden layer. To learn the model a small window moves across the corpus texts and the network tries to predict *the central word of the window* using the words around it.

We assume a window with nine words length and it goes across the unlabeled texts of Hamshahri2 corpus to learn the weights of the network as Persian word embedding vectors. The first and the last four words of each window is assumed to be the input of the network. The central word of the window is assumed to be the label of the output. Thus we have a rich labeled dataset.

Completing the learning process of network weights on all windows of Hamshahri2 corpus, we will have a suitable Persian word embedding set, whose words’ dimension is equal to the size of the hidden layer of the network. The hidden layer size is assumed to be 200 in this work.

The Persian word embedding generated at this stage, maps every words of the Hamshahri2 corpus to a vector in a 200 dimensional vector space. The generated Persian word embedding set contains 300,000 words.

The t-SNE method for visualization can be used to better understand the word embedding environment.

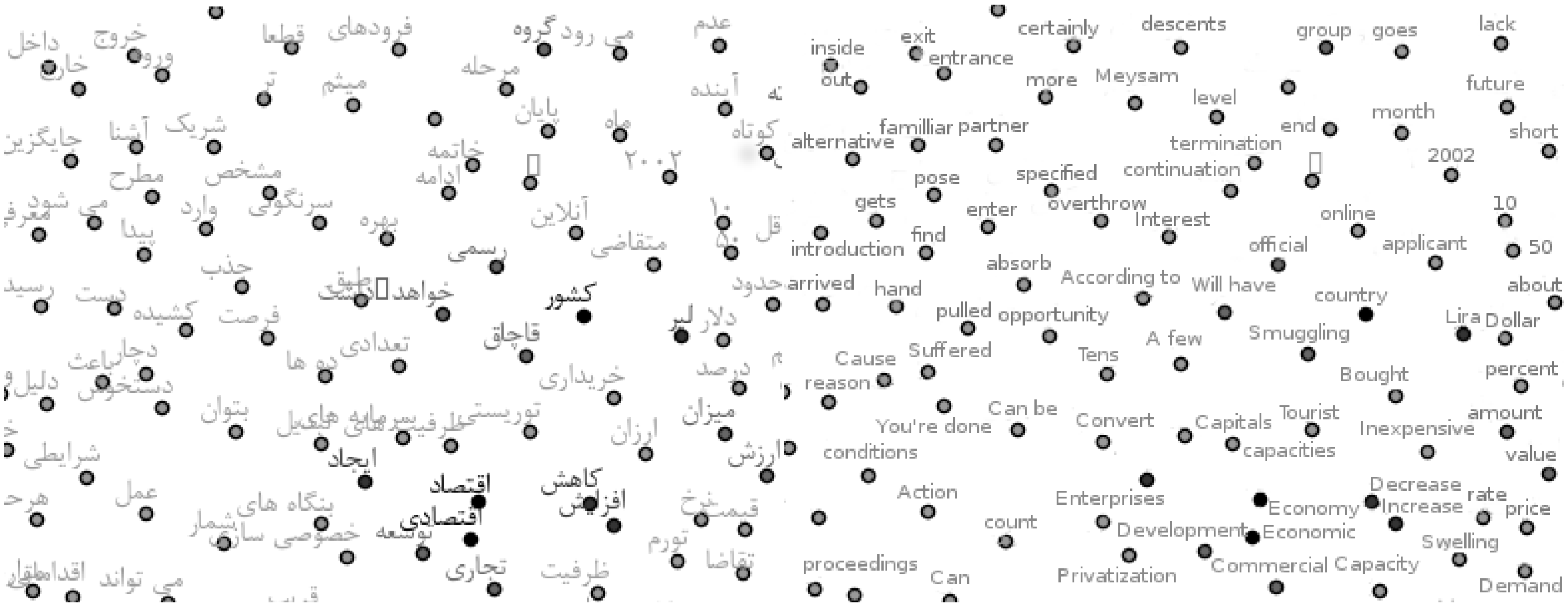

*Figure 2: A Persian word embedding visualization using t-SNE method. Part of the words of one of the texts of the Pasokh corpus visualized in this figure*

In the mapping of the words in  Figure 2: A Persian word embedding visualization using t-SNE method. Part of the words of one of the texts of the Pasokh corpus visualized in this figure, similare words are closer to each other. This issue can also be examined from other dimensions, as another example in  Figure 3: The closest vocabulary to the header terms is given (using the proposed Persian word embedding generated in this work), the closest vocabularies to the header terms is given (using the proposed Persian word embedding generated in this work).

| Isfahan | Semnan | Ahvaz | Mashhad | Darab |
|---|---|---|---|---|
| Shiraz | Zanjan | Abadan | Shiraz | Fasa |
| Tabriz | Yazd | Shiraz | Isfahan | Kazerun |
| Yazd | Qazvin | Tabriz | Tabriz | Firuzabad |
| Mashhad | Hamedan | Khuzestan | Sabzevar | Jahrom |
| kerman | kermanshah | Rasht | Qom | Bavanat |
| Hamedan | Ardabil | Mahshahr | Rasht | Behbahan |
| Zanjan | Lorestan | Sepahan | Tehran | Dashtestan |
| Qazvin | Ilam | Isfahan | Khorasan | Lamerd |
| Kermanshah | Kerman | Omidiyeh | Neyshabur | Estahban |

*Figure 3: The closest vocabulary to the header terms is given (using the proposed Persian word embedding generated in this work)*

In the proposed method, using the relationship between words, the concepts of the input document are represented. In this method, the importance of sentences is determined using semantic and syntactic similarities between words. And Instead of using single words to express concepts, multiple similar words are used. For example, the occurrence of words: computer, keyboard, display, mouse and printer, even though they are not frequently repeated singly in the input document, express a certain concept.

As stated in the introduction, the great reliance of lexical chain based methods on lexical databases is the main weakness of these methods. At this stage, to remove this weakness, an appropriate word embedding for summarization is created that encompasses the semantic and syntactic communication of the words in a broader and more up to date lexical range than that of lexical databases.

The word embedding presented in this work is able to discover relationships present in the outside world that do not exist in common vocabulary databases. For example, this word embedding can detect the relation between the words of Mashhad, Neyshabur and Khorasan. Mashhad is the capital of Khorasan province and Neyshabur is one of the cities of this province Figure 3: The closest vocabulary to the header terms is given (using the proposed Persian word embedding generated in this work). (The common vocabulary databases that can not discover such relationships, are comprehensive lexical databases that carry different meanings for each word along with relationships between them such as: synonyms, antonyms, part of / containing, or more general / more specific relationships. But their construction is manual, costly and time-consuming.)

## 3.3 Extracting the Keywords of the Document

For extracting the keywords of the input document, we first tokenized the words of the document using Hazm tokenizer [40]. [the words of the document tokenized using …] Then we excluded stopwords from input document tokens. The score of each word of the input document calculated using equation (1) [42]:

$$point(w) = TF_{ij} \times IDF_i \quad (1)$$

where w is the intended word, TF calculated from equation (2):

$$TF_{ij} = \frac{f_{ij}}{max_k f_{kj}} \quad (2)$$

where $f_{ij}$ is frequency of the i-th word in the j-th document and $max_k f_{ij}$ is maximum frequency of the words in the input document. The TF is normalized using this division.

Finally, IDF in equation (1) was calculated from equation (3):

$$IDF_i = \log_2(N/n_i) \quad (3)$$

where N is the number of documents of the Hamshahri2 corpus and $n_i$ is the number of documents in the corpus that the i-th word has been observed there.

If a word is not in the Hamshahri2 corpus, there will not be a score for it. Also due to the absence of a vector in the continuous vector space for this word, it is deleted from the decision making cycle.

Therefore, learning word embedding on a richer Persian corpus will cause to increase the accuracy of the method.

## 3.4 Clustering Concepts

In this phase, the concepts present in the input document are constructed using the Persian word embedding obtained in section 3.2. For this purpose:

1. First we sort the keywords of the previous phase according to their calculated scores.
2. Then we map all input document terms into a 300-dimensional space using the prepared Persian word embedding
3. We cluster the concepts of this document into ten different clusters using K-means algorithm:
    - We consider the ten preferred keywords selected in section 3.3 as the primary cluster centers and cluster the entire words of the input document.
    - Each obtained cluster can be considered as a concept. Thus ten key concepts of the document are constructed.
    - Finally, we consider the nearest word to each cluster center as the criterion word for that cluster or concept.
    - The total score of each concept is calculated using the equation (4):

$$point(C) = \sum_{w \in C} (point(w) \times nearness(w)) \tag{4}$$

where w is the word, C is the concept and point(w) is the total score of each word that was calculated based on equation (1).

The nearness(w) indicates the closeness of each word in the intended concept to the concept's criterion word. Therefore the words nearer to the concept's criterion word will have larger linear coefficients and the words farther to that criterion word will have smaller linear coefficients. Thus the nearness of each word to its concept's criterion word affects the final score of the concept. Hence, repetition of more closely situated words in the input document will result in a higher score than repetition of farther words.

## 3.5 Sentence Ranking

For ranking sentences, the following steps are taken:

- first, the input document is read line by line and the sentences of each line are separated using Hazm sentence tokenizer.
- For scoring extracted sentences, equation (5) is used:

$$score(S) = \frac{\sum_{w \in S} point(C)}{N} \quad (5)$$

where S is a sentence, N is its number of words and point(C) is the score of the the intended word's concept.

- By dividing the sentence score into its number of words, we normalized the obtained score, so that shorter and longer sentences would have equal chance of selection.
- Sentences are sorted according to their normalized scores.
- According to the desired summary length, some sentences with the highest score are selected, and are displayed in the order they appear in the document.

# 4 Experimental Results

In this section using ROUGE criterion, our system generated summaries on single-document Pasokh corpus is evaluated and the obtained results are compared with other available Persian summarizers.

## 4.1 Evaluation Measures

ROUGE-N is a measure for evaluation of summarizations [43]. This recall based measure is very close to human evaluation of summaries. This measure calculates the number of common n-grams between the system generated summaries and the reference human made summaries. It's therefore a suitable measure for automatically evaluating summaries produced in all languages. For this work, two public ROUGE evaluation tools are studied:

1. ROUGE: Is a Perl implementation of ROUGE measure that was developed by Mr. C. Lin et al. at the University of Southern California [43]. This implementation dose not support unicode and it generates unrealistic results for the Persian summary evaluation. After obtaining the exaggerated results of this tool for Persian summaries, we realized this great weakness.
2. ROUGE 2: Is a Java implementation of ROUGE-N measure developed by Rxnlp team and is publicly accessible [44]. This tool supports unicode and the obtained results are accurate, but it has only implemented ROUGE-N and not any other variations of ROUGE measure.

In this work, a python implementation of ROUGE-N was developed based on Mr. C. Lin's paper [43]. This tool supports unicode and verifies the results of the ROUGE-2 implementation [44]. According to the above descriptions, the ROUGE-2 is used for summary evaluation in this study.

## 4.2 Pasokh Corpus

Pasokh [37] is a popular corpus for the evaluation of Persian text summarizers. This dataset consists of a large number of Persian news documents on various topics. It contains human-written summaries of the documents in the forms of single-document, multi-document, extractive and abstractive summaries.

The single-document dataset of Pasokh contains 100 Persian news texts that five extractive and five abstractive summaries for each of these news are generated by different human agents.

One hundred news texts of the single-document Pasokh dataset were summarized using the proposed algorithm in this work. The compression ratio of our system summaries was 25 percent. Then we needed to calculate ROUGE-N between each of our system generated summaries and the related 5 Pasokh extractive reference summaries (human-made summaries). For this purpose, ROUGE 2.0 (Java implementation) tool was used, which is mentioned in Evaluation tool section earlier. The average of the 5 ROUGE-N is considered as the evaluation of each of our system summaries. Finally, the average of 100 system summary evaluations was calculated as the final evaluation result.

It should be noted that the news headlines of Pasokh corpus has not been used in summarization process and the results are obtained without taking advantage of headlines.

Pourmasoumi et al. [35] presented Ijaz as an extractive single-document summarizer of Persian news in 2014 which is available online. In this experiment one hundred news texts of the Pasokh corpus were summarized using Ijaz summarizer. The compression ratio was 25 percent, and the results were obtained without using headlines.

The results are reported in Table 1: ROUGE-1 scores (percent) on Pasokh single-document dataset, Table 2: ROUGE-2 scores (percent) on Pasokh single-document dataset and Table 3: ROUGE-3 scores (percent) on Pasokh single-document dataset.

| Systems | ROUGE-1 | | |
|---|---|---|---|
| | Avg_Recall | Avg_Precision | Avg_F-Score |
| Ijaz | 39.3 | 44.8 | 40.5 |
| Shafiee and Shamsfard method [29] | 38.8 | 42.5 | 39.1 |
| Our Proposed Method | **45.4** | **52.4** | **46.8** |

*Table 1: ROUGE-1 scores (percent) on Pasokh single-document dataset*

| Systems | ROUGE-2 | | |
|---|---|---|---|
| | Avg_Recall | Avg_Precision | Avg_F-Score |
| Ijaz | 22.6 | 27.6 | 15.4 |
| Shafiee and Shamsfard method [29] | 21.6 | 24.7 | 22.1 |
| Our Proposed Method | **30.1** | **37.3** | **31.9** |

*Table 2: ROUGE-2 scores (percent) on Pasokh single-document dataset*

| Systems | ROUGE-3 | | |
|---|---|---|---|
| | Avg_Recall | Avg_Precision | Avg_F-Score |
| Ijaz | 18.0 | 22.4 | 19.3 |
| Shafiee and Shamsfard method [29] | 16.7 | 19.3 | 17.1 |
| Our Proposed Method | **26.7** | **34.0** | **28.5** |

*Table 3: ROUGE-3 scores (percent) on Pasokh single-document dataset*

Thus our proposed method in this work has the following advantages over Pourmasoumi et al. method:

- Our proposed method achieves much better results than the proposed method of Pourmasoumi et al. in all ROUGE-1, ROUGE-2 and ROUGE-3 measures.
- The method proposed by Pourmasoumi et al. [35] has taken a supervised learning approach, while our learning approach is unsupervised. As defined by authorities supervised learning requires that the algorithm's possible outputs are already known and that the data used to train the algorithm is already labeled with correct answers. While, unsupervised machine learning is more closely aligned with what some call true artificial intelligence, the idea that a computer can learn to identify complex processes and patterns without a human to provide guidance along the way. Although unsupervised learning is prohibitively complex for some simpler enterprise use cases, it opens the doors to solving problems that humans normally would not tackle.
- Their proposed method is a Persian specific method, while our proposed method can be generalized to other languages.

Shafiee and Shamsfard [29] proposed an approach in extractive single-document Persian summarization in 2017.

Unfortunately, neither their summarizer nor summaries generated by their proposed algorithm are available for comparison, therefore, the algorithm has been implemented.

In this experiment one hundred news texts of the Pasokh corpus were summarized using developed summarizer. The compression ratio was 25 percent, and the results were obtained using headlines. The results are reported in Table 1: ROUGE-1 scores (percent) on Pasokh single-document dataset, Table 2: ROUGE-2 scores (percent) on Pasokh single-document dataset and Table 3: ROUGE-3 scores (percent) on Pasokh single-document dataset.

Our approach has the following advantages over Shafiee and Shamsfard's approach:

- Our proposed method achieves much better results than the "number of similar and related sentences" method of Shafiee and Shamsfard in all ROUGE-1, ROUGE-2 and ROUGE-3 measures.
- Shafiee and Shamsfard's method is supervised, while ours is unsupervised. In order to calculate a feature's weight, they utilize one-third of the Pasokh single-document corpus. To compute a feature's weight, the mean of F-measure scores is calculated to be considered as the final weight of the selected feature for single-document summarization.

- Their proposed method depends on enriching and keeping up to date the FarsNet lexical database, that is very costly and time consuming, while our method depends on unsupervised learning of the target language word embedding.
- Their proposed method is a Persian specific method, while our proposed method can be generalized to other languages.
- Their method has used the news headlines in the summarization process, while our method has obtained the results without using headlines.

Hassel and Mazdak created FarsiSum [25] in 2004 as one of the first Persian text summarizers reported in related literature. The available version of FarsiSum summarizer in their website has a number of bugs. For example, the length of the summary FarsiSum produces has a significant difference with the requested compression ratio percentage. According to previous studies [29], [35], the results of our proposed method on Pasokh corpus are much higher than the results obtained by FarsiSum summarizer.

# 5 Conclusion

In this paper, a novel method of extractive generic document summarization based on perceiving the concepts present in sentences is proposed. Therefore after unsupervised learning of the target language word embedding, input document concepts are clustered based on the learned word feature vectors (hence the proposed method can be generalized to other languages). After allocating scores to each conceptual cluster, sentences are ranked and selected based on the significance of the concepts present in each sentence.

One of the most important challenges in recent researches in the field of summarizing Persian texts is the lack of a rich lexical database in Persian language that can be used to measure semantic similarities. In this research, by constructing a Persian word embedding using Hamshahri2 corpus, we were able to correctly answer this shortage and provide a new method for summarizing the texts according to the semantic and syntactic relations learned.

Using the relationship between words, the concepts discussed in the input document are represented. In this method, the importance of sentences is determined using semantic and syntactic similarities between words. Instead of using single words to express concepts, different related words are used. We evaluated the proposed method on Pasokh single-document dataset using the ROUGE evaluation mesure. Without using any hand-crafted features, our proposed method achieves state-of-the-art results. For system summaries generated with 25 percent compression ratio on Pasokh single-document corpus, ROUGE-1, ROUGE-2 and ROUGE-3 recall scores were 45, 30 and 27 percent, respectively.

Evaluation of our proposed method for summarization of other languages is suggested for future works. Learning word embedding on richer Persian corpuses may be effective in increasing the accuracy of our method. using PageRank algorithm to produce the concept similarity graph and to find more significant concepts may also increase the accuracy of our concept selection algorithm. Using exploited MMR (Maximum Marginal Relevance) greedy algorithm in sentence selection process may decrease the redundancy of the selected sentences in our proposed method.